\documentclass[journal]{IEEEtran}

\ifCLASSINFOpdf
\else
\fi
\usepackage{graphicx}
\usepackage{epstopdf}
\usepackage{ tipa }
\usepackage{subfigure}
\usepackage{amsmath}
\usepackage{amsmath,amssymb}
\usepackage{lmodern}
\usepackage{iftex}
\usepackage{color}
\usepackage{amsmath,amssymb}
\usepackage{lmodern}
\usepackage{longtable}
\usepackage{float}   
\usepackage{booktabs} 
\usepackage{makecell}
\usepackage{hyperref}
\usepackage{tabularray}
\usepackage{float}
\usepackage{cite}

\usepackage{booktabs}
\usepackage{colortbl}

\begin{document}
%
\title{A Transformer-based Prediction Method for Depth of Anesthesia During Target-controlled Infusion of Propofol and Remifentanil}
%
%
%
%

\author{Yongkang~He,
        Siyuan~Peng,~\IEEEmembership{Member,~IEEE,}
        Mingjin~Chen,
        Zhijing~Yang,
        Yuanhui~Chen
 \thanks{Y. He, S. Peng, M. Chen and Z. Yang are with the School of Information Engineering, Guangdong University of Technology, 510006, China (e-mail: 1181309500@qq.com, peng0074@gdut.edu.cn, 2112103033@mail2.gdut.edu.cn, yzhj@gdut.edu.cn).}
\thanks{Y. Chen is with the Zhejiang Hospital of Integrated Traditional Chinese and Western Medicine, 310005, China (e-mail: chenyuanhui\_831@hotmail.com).}
}

\maketitle

\begin{abstract}

Accurately predicting anesthetic effects is essential for target-controlled infusion systems. The traditional (PK-PD) models for Bispectral index (BIS) prediction require manual selection of model parameters, which can be challenging in clinical settings. Recently proposed deep learning methods can only capture general trends and may not predict abrupt changes in BIS. To address these issues, we propose a transformer-based method for predicting the depth of anesthesia (DOA) using drug infusions of propofol and remifentanil. Our method employs long short-term memory (LSTM) and gate residual network (GRN) networks to improve the efficiency of feature fusion and applies an attention mechanism to discover the interactions between the drugs. We also use label distribution smoothing and reweighting losses to address data imbalance. Experimental results show that our proposed method outperforms traditional PK-PD models and previous deep learning methods, effectively predicting anesthetic depth under sudden and deep anesthesia conditions.

\end{abstract}


\IEEEpeerreviewmaketitle

\section{Introduction}
\IEEEPARstart{W}{ith} the advancement of automated control technology, intravenous target-controlled infusion techniques are increasingly being utilized in anesthesia procedures \cite{anderson2019practicalities,white1990intravenous}. The pharmacokinetic-pharmacodynamic (PK-PD) model \cite{egan1995remifentanil,white1988propofol} is currently widely adopted in infusion pumps to calculate the effector compartment concentration of anesthetic drugs. However, the traditional PK-PD model has a significant limitation. In clinical practice, it typically requires the selection of multiple parameters due to individual organism differences \cite{minto2008contributions}. This is because the drug effect between the drug dose and a specific organism is unclear. Even if the same dose of an anesthetic drug is administered at the same time, physiological responses vary from person to person \cite{schuttler2000population}. To date, no reasonable or effective research has been conducted on precisely administering drugs according to a specific individual, while combining with the existing PK-PD model.

An accurate drug efficacy prediction model is essential for intravenous target-controlled infusion systems \cite{merigo2019optimized, van2019optimizing}. In recent years, deep learning methods have been investigated for addressing this problem \cite{connor2019artificial, caelena2011real}. Compared to traditional PK-PD prediction models, deep learning methods have the advantage of complex nonlinear dynamic computation, resulting in good prediction performance under different situations such as complex environmental information, unclear knowledge background, and unclear inference rules. Lee  \emph{et al.} proposed a method in \cite{lee2018prediction} that combines the PK-PD model framework with the long short-term memory (LSTM) network to extract features from drug injection history information, and then incorporates human physiological characteristics such as age, gender, height, and weight to predict the Bispectral Index (BIS). Although this method shows significant improvement in anesthesia depth prediction compared to previous PK-PD-based methods, it performs poorly on samples with large fluctuations in BIS. As a result, the deep learning-based prediction method proposed in \cite{lee2018prediction} is less efficient at predicting the depth of anesthesia (DOA) during unexpected situations.

\begin{figure*}[htbp]
\centering
\includegraphics[width=7.0in, height=3.0in]{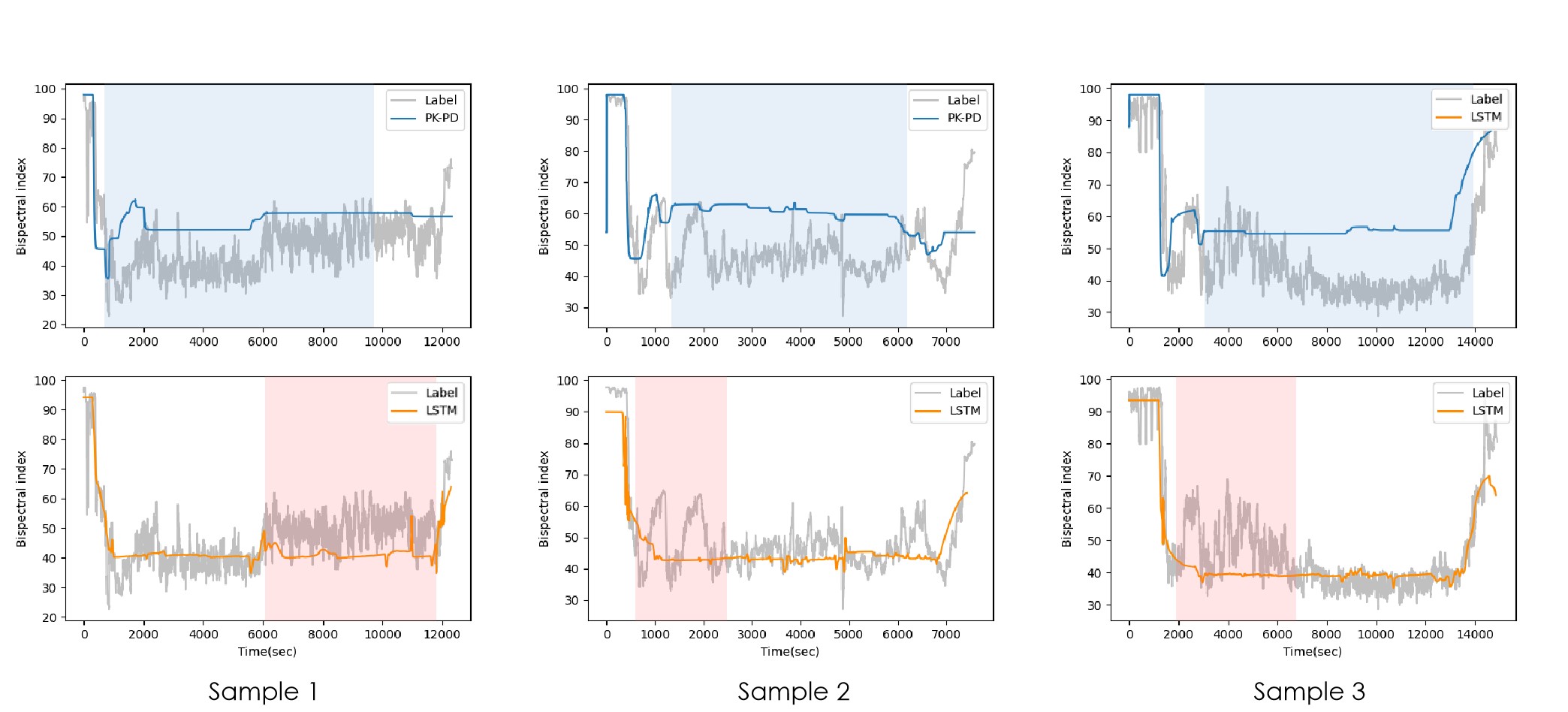}
\caption{Prediction results of PK-PD and LSTM for different samples. Top: the blue line is the predicted values of the PK-PD model, which shows a great deviation from the ground true value, especially in the light blue area. Bottom: the orange line is the predicted values of the LSTM model, which usually has relatively poor results (see the light pink area) under the conditions of abrupt change.}
\label{pk-pd and lstm compare}
\end{figure*}

In addition, some researchers have successfully utilized the Electroencephalogram (EEG) signal \cite{wang2021eye} to calculate the BIS value. For example, Li \emph{et al.} used the Butterworth filter to extract several features such as column entropy, sample entropy, wavelet entropy, and band power from EEG, and then input these features to the sparse denoising autoencoder and long short term memory (SDAE-LSTM) network to predict the DOA \cite{li2020monitoring}. Combining the signal processing and deep learning technique, this method has high prediction accuracy for the DOA.
However, the EEG-based prediction method is less practical than the PK-PD-based prediction method, since it requires the huge amount of the EEG signal data and is easily interfered by the electromagnetic. Furthermore, from the viewpoint of real-world applications, the proposed method in \cite{li2020monitoring} is hard to directly establish a simulation environment for EEG signals in the field of anesthesia control.

\begin{figure}[htbp]
\centering
\includegraphics[scale=0.15]{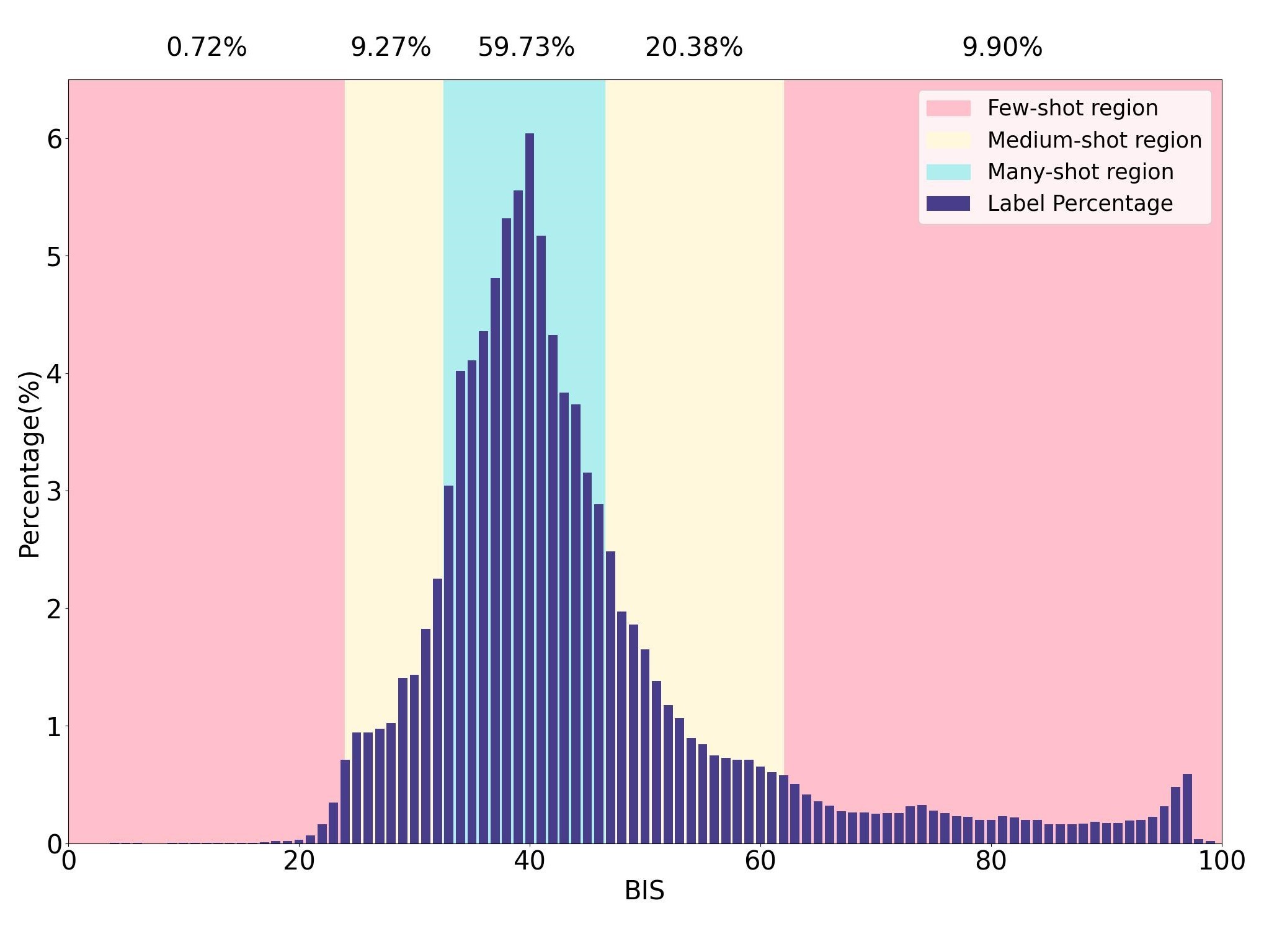} \vskip -0.5cm
\caption{Label distribution of the dataset, which is
divided into three regions including the many-shot region (59.73\%), the medium-shot region (29.65\%), and the few-shot region (10.62\%).}
\label{label distribution}
\end{figure}

Fig. \ref{pk-pd and lstm compare} shows the prediction results of the PK-PD model \cite{egan1995remifentanil} and the LSTM-based deep learning method \cite{lee2018prediction} for different samples. The figure reveals two main drawbacks of previous approaches to predicting the depth of anesthesia (DOA). Firstly, when sudden changes in bispectral index (BIS) occur during the maintenance period, the prediction results of previous methods remain relatively stable and do not reflect the actual changes, as can be seen in the light blue and pink areas in Fig. \ref{pk-pd and lstm compare}. Secondly, the BIS data collected for anesthesia clinical records are often unbalanced, with most values falling in the $30-50$ range, as illustrated in Fig. \ref{label distribution}. Previous works have neglected the few-shot region, leading to overfitting in many-shot regions and inaccurate predictions in other regions.

This paper proposes a new deep learning method, based on transformer architecture, to accurately predict the DOA using the drug infusion of propofol and remifentanil. The proposed method uses the fusion of human parameters, drug injection history, and derived multimodal features to enhance prediction accuracy. The PK-PD model is embedded at the beginning of the network to provide pseudo-historical information, which is corrected in the training phase using the LSTM network and bottleneck layer. The gate residual network (GRN) module is then applied to fuse multidimensional features and patient context information, suppress irrelevant variables, and aggregate physiological characteristics into each time step. An improved attention mechanism is used to learn long-term dependencies among mixed features for exploiting drug-drug interactions. To overcome data imbalance, the proposed method uses label distribution smoothing and reweighting losses to prevent overfitting in many-shot regions and exhibit good prediction ability in other regions.

In summary, the main contribution of this work are as follows:
\begin{enumerate}
  \item A new transformer-based deep learning framework is proposed to predict the DOA by using the drug infusion history of propofol and remifentanil simultaneously, which can overcome the limitations of previous DOA prediction methods;
  \item A feature fusion layer is developed in the proposed method to combine dynamic and static information from different modalities to achieve the fusion of temporal and textual information, enabling the entire network to fully consider the response of patients with different ages, genders, heights and weights for the same drug;
  \item The label distribution smoothing and reweighting losses are used to solve the issue of data imbalance in different intervals in the filed of DOA.
\end{enumerate}

The remainder of this paper is organized as follows. The related works are introduced in Section II. The proposed method is presented in Section III. The experimental results are shown in Section IV. Finally, the conclusion is given in Section V.
\section{Related Works}
\subsection{Prediction of Anesthetic Efficacy}

Anesthetic prediction methods based on PK-PD models have been widely used in clinical drug effect prediction \cite{minto2008contributions,short2016refining}. These methods model the transfer and metabolism of drugs in each component of the human body by solving a system of differential equations. However, PK-PD models with fixed parameters often have poor performance due to inter-patient variability. Although an optimization approach in \cite{gonzalez2020adaptive} was used to identify the parameters for different patients, it still required measuring BIS values during the procedure for optimization. Recently, deep learning methods have been proposed for drug effect prediction based on time series prediction \cite{caelena2011real, lee2018prediction}. For instance, in \cite{lee2018prediction}, an LSTM model was used to extract long-term and short-term memory of drug injection records, and combined with patient characteristics for BIS value prediction. However, the proposed method did not pay enough attention to sparse data samples (e.g. deep anesthesia states with BIS below 40), which often leads to poor performance on imbalanced data.

\subsection{Multivariate Time Series Forecasting}
Distinguishing from the univariate time series forecasting, the distinction between variables needs to be considered under the task of multivariate input, since different variables may be heterogeneous or even belong to different modalities. Previous studies have often considered in the location of the feature fusion layer. For example, Anastasopoulos \emph{et al.} experimentally verified that multimodal data fusion usually have better performance in the middle or deep layers of the network \cite{anastasopoulos2019neural}. Pérez-Rúa \emph{et al.} discovered the optimal architecture from a large number of possible combinations of positions of the network by means of a sequential model-based exploration method approach \cite{perez2019mfas}. In recent years, heterogeneous feature fusion based on the attention mechanisms \cite{vaswani2017attention} technique has gradually applied in the practical tasks \cite{lim2021temporal,arevalo2017gated}. Bryan Lim \emph{et al.} proposed to use a gating mechanism and an interpretable attention mechanism to achieve multilevel time series prediction \cite{lim2021temporal}. Arevalo \emph{et al.} developed a gated multimodal units, which found the best combination from different combinations of data and allowed to apply this fusion strategy anywhere in the model \cite{arevalo2017gated}.
\subsection{Data Imbalance}
For the anesthesia clinical records, the collected data are often unbalanced (as shown in the Fig.\ref{label distribution}, most of the BIS values are in the range of 30-50). To solve this issue, the proposed method in \cite{garcia2009evolutionary} adjusted the distribution among the data by resampling the samples. However, such methods are often difficult to grasp the sampling ratio, leading to oversampling easily. In order to further address this problem of data imbalance, some improved methods have been successfully proposed in \cite{cao2019learning,dong2018imbalanced}, which adjusted the learning weights of the loss function for a small number of samples. According to \cite{liu2019large}, one can observe that the methods that directly adjust the weights of the loss function requires a high degree of differentiation between categories. Hence, in \cite{yang2021delving}, a novel method has been developed to improve the performance of reweighting loss by smoothing the label distribution of the samples.
\section{Methodology}
\begin{figure*}[htbp]
\centering
\includegraphics[scale=0.50]{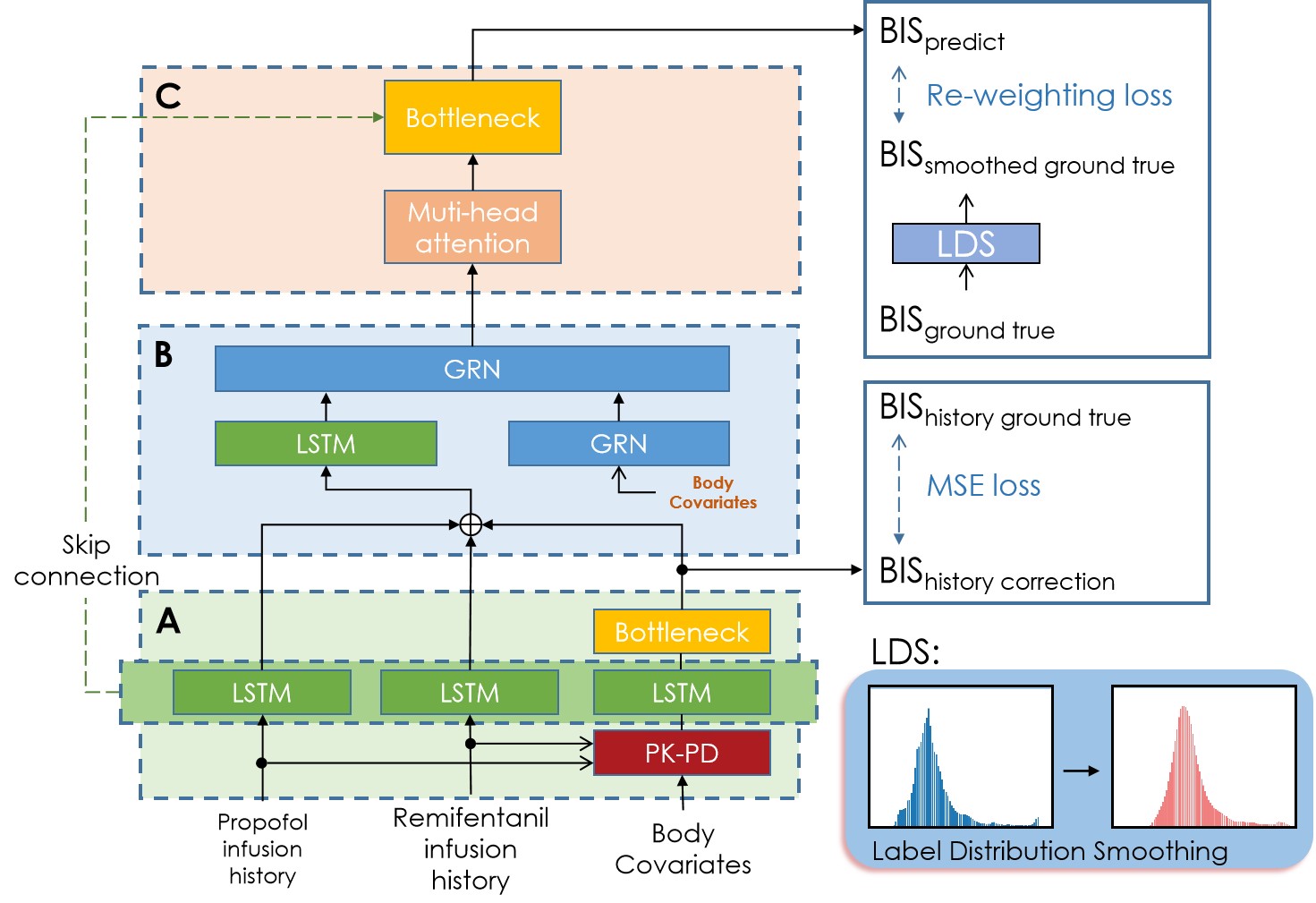} \vskip -0.2cm
\caption{An overview of our model. Our framework consists of three components, A: Pharmacodynamic encoder, it is used to extract drug history information. Right side of A: correction of pseudo-bis values calculated by PK-PD model; B: Feature fusion layer, it combines the dynamic temporal information with the static human physiological features; C: Temporal fusion decoder, it is adopted to learn the long-term dependencies of temporal hybrid features using a multi-headed attention mechanism.}
\label{model architecture}
\end{figure*}
\subsection{Problem definition}
In this paper, our goal is to design a transformer-based network for accurate drug effect prediction by using the drug injection history of propofol and remifentanil and body covariates together. The overall framework of our proposed method are shown in Fig. \ref{model architecture}, which consists of three components:  A) \textbf{drug effect encoder}, it is used for extracting the temporal features from the drug injection history, B) \textbf{feature fusion layer}, it is used for fusing different dynamic and static information, and C)  \textbf{temporal fusion decoder}, it is used for learning the mixed features of different long-term dependencies. A re-weighted root mean square loss function is adopted in the training phase to overcome the drawback of data imbalance. First, the two anesthetic drugs that are often used in combination in anesthetic surgery are described in detail in the next subsection.

\subsection{Anesthesia clinical record and feature extraction}
\subsubsection{Prorofol infusion history}
Propofol is a widely used anesthetic drug in general anesthesia procedures \cite{bryson1995propofol}, which provides rapid and stable hypnosis function and has additional or synergistic hypnotic effects with other drugs used in anesthesia (such as barbiturates, benzodiazepines, opioids and ketamine) \cite{struys2000comparison}. Thanks to its large absorption and rapid elimination by the body, propofol has become the best anesthetic target-controlled infusion (TCI) drug. In the automated target-controlled infusion systems, the injection rate of propofol is often used as one of the most important characteristics for calculating the BIS prediction values \cite{short2016refining}. In our work, the injection history of propofol in the range of 1800s before $t$ is adopted as a model feature to predict the BIS value at moment $t+1$.

\subsubsection{Remifentanil infusion history}

Remifentanil is commonly used as a supplement for the general anesthesia and is extensively metabolized extrahepatically by blood and tissue non-specific esterases, resulting in an extremely rapid clearance efficiency (3 L/min) \cite{egan1995remifentanil}. However, when the synergistic effect of remifentanil and propofol is given in \cite{gonzalez2020adaptive}, similarly, the injection history of remifentanil at 1800s before moment $t$ into the proposed model is utilized to predict the BIS value at moment $t+1$.

\subsubsection{Drug effect site concentration}

The concentration of the drug effect represents the ideal concentration of the drug at the site of action in the body. To some extent, it reflects the effect of the anesthetic drug on the DOA. Note that the effector compartment is not real and therefore cannot be measured directly. Based on the traditional three-compartment model \cite{minto2008contributions}, it can be calculated by a pharmacokinetic model. Combining the effector compartment concentration and the pharmacokinetic model, the simple pseudo-BIS values are able to initially calculate. Then the neural network method is applied to correct the pseudo-BIS values calculated by the PK-PD model to briefly provide the historical information for the proposed model, which is illustrated in part A of Fig. \ref{model architecture}.

\subsection{Model architecture}

Our proposed method combines a recurrent neural network (RNN) with an attention mechanism and transformer architecture. RNNs are effective at capturing features from time series data and preserving memory, but using a single RNN can be difficult when dealing with multiple, heterogeneous inputs. Simply combining different types of data in the network can lead to the loss of important characteristics \cite{perez2019mfas}. To address this issue, our method extracts features from different inputs using separate long short-term memory (LSTM) modules and combines them using a feature fusion layer that controls the mixing of multiple types of information. Furthermore, we incorporate static covariates at each time step to explore the relationship between dynamic and static information.
To compensate for the uneven distribution of data caused by the small number of samples, we use the label distribution smoothing method to smooth data with highly unbalanced label categories. At the end of the model, we use a reweighting loss function to assign higher loss weights to categories with sparse numbers (such as deep anesthesia states with BIS values between 20 and 30). This encourages the network to pay more attention to deep anesthesia states with critical sample sizes, despite their small representation in the data.

Our model consists of the following three parts:
\subsubsection{Drug effect encoder}

\begin{figure}[htbp]
\centering
\includegraphics[scale=0.45]{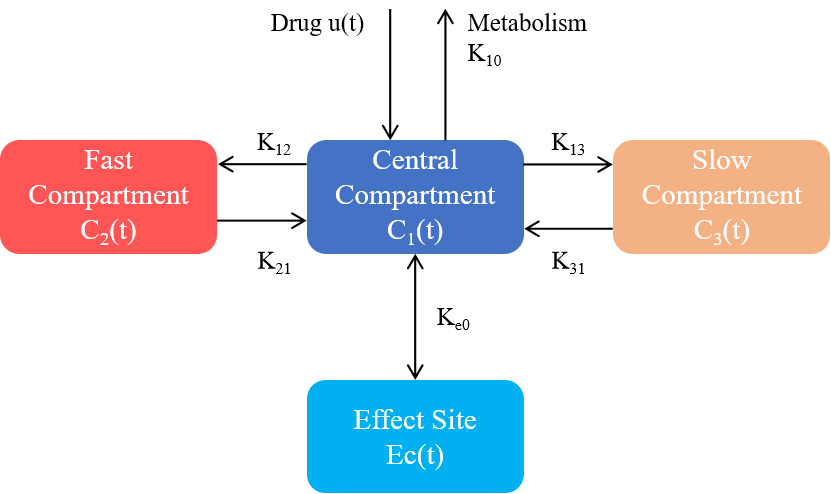}
\caption{Structure of the PK model based on three compartments and an effect site compartment.}
\label{pkpd model}
\end{figure}

To capture pharmacological feedback under different anesthesia stages, the proposed network first applies the PK-PD model to calculate the pseudo historical information of BIS, and then uses LSTM and bottleneck (as shown on the right side of part A in Fig. \ref{model architecture}) to correct the pseudo historical information. The PK-PD model is a classical model widely used in anesthesiology to calculate the effects of anesthetic drugs. The PK-PD model uses a three-compartment model that utilizes the drug infusion rate as input to simulate the transfer and metabolism of the drug between various regions of the body, and reflects the drug effect with an ideal effector compartment. Assuming that the drug effect with an ideal effector compartment has a negligible volume, the clinical effect of the drug is quantified as the effector concentration $E_c$. Fig. \ref{pkpd model} shows the general structure of the PK model used in anesthesia, where the central chamber represents the plasma and tissue, and the fast and slow chambers represent the peripheral chambers, including less perfused organs \cite{blusse2019population}.

According to this structure, the Schnider and Minto model \cite{minto2008contributions} parameters is used to calculate the effect  concentration \( E_{c} \) by solving the system of differential equations. Then the response surface model proposed by Short et al. \cite{short2016refining} is adopted to calculate the BIS values:
\begin{small}
\begin{equation}
B I S=B I S_{0}+\left(B I S_{\min }-B I S_{0}\right) \frac{\left(\frac{E c_{r}}{E c_{50 r}}+\frac{E c_{p}}{E c_{50 p}}\right)^{\gamma}}{1+\left(\frac{E c_{r}}{E c_{50 r}}+\frac{E c_{p}}{E c_{50 p}}\right)^{\gamma}}
\end{equation}
\end{small}
where $Ec_{r}$ and $Ec_{p}$ are the effect concentrations of propofol and remifentanil respectively. $\gamma$ stands for the nonlinear shape of the sigmoid curve. $Ec_{50r}$ and $Ec_{50p}$ denotes the effector site concentration corresponding to 50\% of the maximum clinical effect, which can calculated from Eqn.(\ref{pkpd-e1}) to Eqn.(\ref{pkpd-e4}):

\begin{equation}
\begin{gathered}
V_{1} \frac{d C_{1}(t)}{d t}=V_{2} C_{2}(t) k_{21}+V_{3} C_{3}(t) k_{31} \\
-V_{1} C_{1}(t)\left(k_{10}+k_{12}+k_{13}\right)+u(t)
\end{gathered}
\label{pkpd-e1}
\end{equation}

\begin{equation}
V_{2} \frac{d C_{2}(t)}{d t}=V_{1} C_{1}(t) k_{12}+V_{2} C_{2}(t) k_{21}
\end{equation}

\begin{equation}
V_{3} \frac{d C_{3}(t)}{d t}=V_{1} C_{1}(t) k_{13}+V_{3} C_{3}(t) k_{31}
\end{equation}

\begin{equation}
\frac{d E c(t)}{d t}=C_{1}(t) k e_{0}-C_{e}(t) k e_{0}
\label{pkpd-e4}
\end{equation}where \( V_{1},V_{2},V_{3} \) are the volumes of central compartment, fast compartment and slow compartment, respectively, and \( k_{ij} \) is the rate of drug transfer between the chambers. All of them are calculated from human physiological characteristics (age, sex, height, weight), as shown in Table \ref{pk-pd params}.


However, a PK model with a limited number of compartments and covariates may be insufficient to accurately account for propofol kinetics. Therefore, we used a separate LSTM modules and Bottleneck (a three fully connected layer) to increase PK-PD model’s parameter quantity by extract features from pseudo-history BIS values and injection history of propofol and remifentanil. When encountering a population that is very different from our training set, we can fine-tune its parameters by using a small number of samples to retrain the model.

Because different drugs have different elimination times, For example, propofol takes longer to be absorbed (elimination of 66\% in about 25 minutes) while remifentanil has a faster absorption and elimination. Therefore, we used three independent LSTM modules with different parameters to address this issue.

\subsubsection{Feature fusion layer}
It should be noted that the effects of drugs can vary among different populations, even when administered at the same dose and time \cite{schuttler2000population}. Generally, the variability between patients makes it impossible to overlook their physiological characteristics. However, the relationship between a patient's physiological information and the corresponding drug effects is not always direct, and this can negatively affect the results. As a result, using a simple fully connected layer may degrade the performance of the model. In our proposed method, we use the gate residual network (GRN) from \cite{lim2021temporal} to control the input variables. GRN uses a gating mechanism to eliminate irrelevant noisy variables and extract important parts from variables in multivariate regression, where the specific contributions of variables to the output are often unknown. The human physiological information (age, gender, height, and weight) is integrated into the network based on GRN to combine temporal and static information features.
\begin{figure}[htbp]
\centering
\includegraphics[scale=0.45]{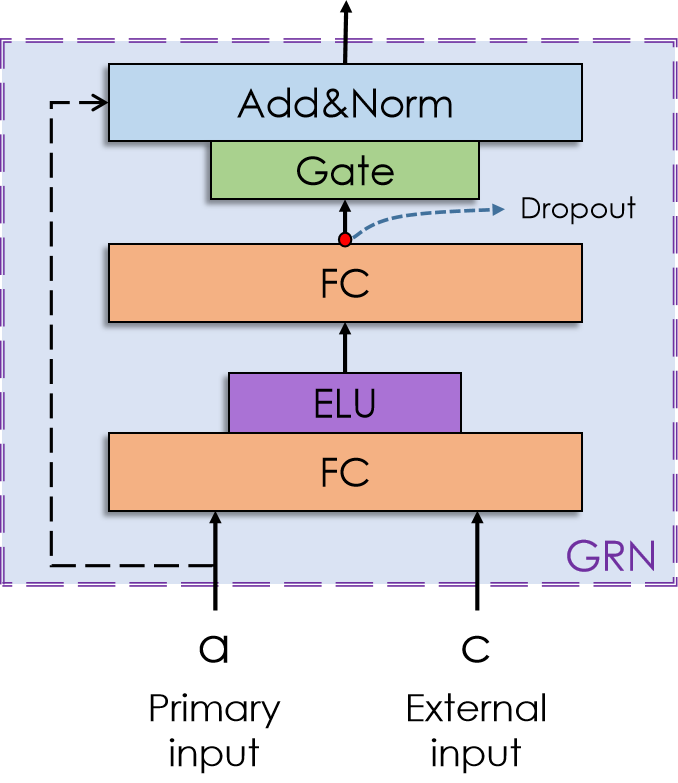}
\caption{Structure of GRN. The gated residual network blocks enable the efficient information flow with the skip connections and the gating layer.}
\label{GRN}
\end{figure}

Fig. \ref{GRN} demonstrates the structure of GRN, in which the parameters \(\boldsymbol{a}\) and \(\boldsymbol{c}\) denote the primary input and external input of GRN respectively. Specific to our method, \(\boldsymbol{a}\) is generated by drug effect encoder, which is temporal feature about drug infusion history and BIS pseudo history. And \(\boldsymbol{c}\) is patient physiological information.
After first fully connection (FC) layer, Exponential Linear Unit (ELU) activation function be used to accelerates the learn speed by reducing the effect of bias offset and making the normal gradient closer to the unit natural gradient \cite{clevert2015fast}. The gated linear unit (GLU) is applied at the output layer to select the input information, which can effectively suppress the noisy information that is not relevant to the output result \cite{dauphin2017language}. Finally, the primary input \(\boldsymbol{a}\) is connected to the output layer after passing through the Layernorm Layer \cite{ba2016layer}, which serves to normalize a single sample and accelerate the convergence of the entire network.
\subsubsection{Temporal fusion decoder}
In our work, the interpretable multi-headed attention mechanism is used to learn long-term and short-term dependencies between different time steps from the multidimensional features with a mixture of temporal and static information. Different from the classical multi-headed attention, the interpretable multi-headed attention modifies the calculation of the attention weights in multiple heads, for enhancing the characterization of specific features by:
\begin{equation}
\mathcal{F}(\boldsymbol{Q}, \boldsymbol{K}, \boldsymbol{V})=\tilde{\boldsymbol{H}} \boldsymbol{W}_{\boldsymbol{H}}
\end{equation}
where $\boldsymbol{Q}$, $\boldsymbol{K}$, and $\boldsymbol{V}$ denote the keys, queries and values respectively, all of them are the vectors and originated from the input features. $\boldsymbol{W}_{\boldsymbol{H}}$ denotes the weights for $\tilde{\boldsymbol{H}}$, which is used for linear mapping. In addition, $\tilde{\boldsymbol{H}}$ is obtained by summing each head:

\begin{equation}
\begin{aligned}
\tilde{\boldsymbol{H}} &=\tilde{A}(\boldsymbol{Q}, \boldsymbol{K}) \boldsymbol{V} \boldsymbol{W}_{V} \\
&=\left\{\frac{1}{m_{H}} \sum_{h=1}^{m_{H}} A\left(\boldsymbol{Q} \boldsymbol{W}_{Q}^{(h)}, \boldsymbol{K} \boldsymbol{W}_{K}^{(h)}\right)\right\} \boldsymbol{V} \boldsymbol{W}_{V} \\
&=\frac{1}{m_{H}} \sum_{h=1}^{m_{H}} \text { Attention }\left(\boldsymbol{Q} \boldsymbol{W}_{Q}^{(h)}, \boldsymbol{K} \boldsymbol{W}_{K}^{(h)}, \boldsymbol{V} \boldsymbol{W}_{V}\right)
\end{aligned}
\end{equation}
where $\boldsymbol{W}_{Q}^{(h)}$ and $\boldsymbol{W}_{K}^{(h)}$ are the head-specific weights for $\boldsymbol{Q}$ and $\boldsymbol{K}$, and $\boldsymbol{W}_{V}$ are the value weights shared across all heads.

Specifically, for the temporal fusion features \( Z(t)=[z(t-k), \cdots, z(t)]^{T} \) obtained by the feature fusion layer, the multi-headed attention mechanism is applied:
\begin{equation}
\boldsymbol{A}(t)=\mathcal{F}(Z(t), Z(t), Z(t))
\end{equation}
to yield \( B (t)=[\beta(t-k), \cdots, \beta (t)]^{T} \). Finally, the last time step \( \beta (t) \) of \( B (t) \) is utilized as the output, and then the skip connection is adopted to combine the hidden states \(h_{\tau}^{(1)}, h_{\tau}^{(2)}, h_{\tau}^{(3)}\) of the last layer of the three LSTM modules in the encoder and \(\beta (t)\) into the bottleneck. The output is the final BIS prediction values. The skip connections are used to facilitate feature fusion as well as to prevent performance degradation caused by over-deepening the network.
\begin{table*}[]
\renewcommand\arraystretch{1.5}
\caption{\textbf{PATIENT CHARACTERISTICS, MEAN±STANDARD DEVIATION (MIN-MAX)}}
\centering
\setlength{\tabcolsep}{4.5mm}{
\begin{tabular}{lccc}
\toprule
 &Training Data Set &Validation Data Set&Testing Data Set \\
\midrule
N & 180 & 76 & 76 \\
Age(yr) & 56.1 ± 14.0 (17-82) & 56.3 ± 15.0 (17-79) &
56.2 ± 15.1 (17-79) \\
Sex(male/female) & 113/67 & 47/29 & 40/36 \\
weight(kg) & 61.5 ± 10.2 (37.9-98.1) & 60.7 ± 10.3 (37.9-98.1) & 60.0 ± 9.8 (37.9-81.6) \\
Height(cm) & 163.2 ± 8.2 (138.8-186.6) & 162.3 ± 7.9 (138.8-182.0) &
161.2 ± 7.5 (138.8-182.0) \\
Median BIS & 41.1 ± 5.4 (25.9-59.5) & 43.1 ± 6.1 (23.1-57.2) & 42.5 ± 5.8 (30.6-55.9) \\
Propofol total dose($g$) & 1.19 ± 0.63 (0.28-3.41) & 1.27 ± 0.71 (0.32-3.31) &
1.32 ± 0.71 (0.30-4.24) \\
Propofol Median Ce($\mu g/ml$) & 3.02 ± 0.47 (1.91-4.30) & 3.06 ± 0.49 (2.00-4.00) & 3.05 ± 0.50 (1.60-4.00) \\
Remifientanil total dose($g$) & 1.46 ± 1.01 (0.29-6.29) & 1.43 ± 0.84 (0.25-3.70) &
1.46 ± 0.91 (0.34-5.16) \\
Remifientanil Median Ce($\mu g/ml$) & 3.73 ± 1.08 (1.50-6.01) & 3.67 ± 0.95 (2.00-6.97) &
3.70 ± 0.87 (2.00-6.00) \\
\bottomrule
\end{tabular}}
\label{data stastic}
\end{table*}

\begin{table*}
\renewcommand\arraystretch{1.6}
\caption{\textbf{PK and PD parameters of propofol and remifentanil.}}
\centering
\setlength{\tabcolsep}{7mm}{
\begin{tabular}{lcc}
\toprule
Params &Propofol&Remifentanil  \\
\midrule
model &Schnider &Minto \\
$V_{1}$& 4.27 & \makecell{5.1 - 0.0201*(age - 40)\\ + 0.072*(lbm - 55)} \\
$V_{2}$& 18.9 - 0.391*(age - 53) & 9.82 - 0.0811*(age - 40) \\
$V_{3}$& 238 & 5.42 \\

$C_{1}$& \makecell{1.89 + (wgt - 77)*0.0456 \\- (lbm - 59)*0.0681 + (hgt - 18)*0.0264} &  \makecell{2.6 - 0.0162*(age - 40)\\ + 0.0191*(lbm - 55)} \\
$C_{2}$& 1.29 - 0.024*(age - 53) & 2.05 - 0.0301*(age - 40)  \\
$C_{3}$& 0.836  & 0.076 - 0.00113*(age - 40)  \\
$ke_{0}$ &0.46 & 0.595 - 0.007*(age - 40)\\

$E_{0} - E_{max}(BIS)$  & 98-0  & 98-0  \\
$Ec_{50} (\mu g/mL)$  & 4.47 & 19.3  \\
$\gamma$  & 1.43 & 1.43 \\
\bottomrule
\multicolumn{3}{l}{\small  \makecell[l]{$\bullet$
$Age = age(y); wgt = weight(kg); hgt = height (cm); lbm = lean \; body \; mass$ \\
$ lbm_{male} = 1.1*wgt - 128*(wgt/hgt)^{2},
lbm_{female} = 1.07*wgt - 140*(wgt/hgt)^{2}$
}}\\

\end{tabular}}
\label{pk-pd params}
\end{table*}

\subsubsection{Label distribution smoothing}
In classification tasks, it is popular to increase the loss weights of samples from a few categories. This can lead the used network to focus on those categories with less data for solving the issue of data imbalance \cite{liu2019large}. However, weighting of the loss function usually requires a high correlation between the label distribution of the samples and the error distribution. Furthermore, a dataset with a continuous label space usually has the following properties: the error distribution is smooth and no longer correlates well with the label density distribution. Therefore, to address the imbalance of the label distribution in our task, the label distribution smoothing proposed in \cite{yang2021delving} is utilized, which adopts a Gaussian kernel function convolved with the empirical density of the labels to extract a kernel-smoothed new label distribution, given by:
\begin{equation}
\tilde{p}\left(y^{\prime}\right) \triangleq \int_{\mathcal{Y}} \mathrm{k}\left(y, y^{\prime}\right) p(y) d y
\end{equation}
where $p(y)$ is the number of appearances of label y in the training data, $\tilde{p}\left(y^{\prime}\right)$ is the effective density of label  $y^{\prime}$ , and $k(y, y^{\prime})$ is Gaussian kernel, which characterizes the similarity between target value $y^{\prime}$ and any $y$ in the target space. The new distribution has an excellent negative Pearson correlation with the error. After obtaining a more efficient label density, the usual methods are able to apply for solving the label imbalance in our task. The details are described in the following section.
\subsubsection{Loss function}
We first define some notations. We parameterize the encoder (part A in Fig. \ref{model architecture}) as $\theta_{en}$ (excluding the fixed parameters in PK-PD), the feature fusion layer (part B) is denoted as $\theta_{fl}$, and the decoder (part C) is denoted as $\theta_{de}$.
Suppose we are predicting BIS at moment $\tau$, the pseudo-history BIS of PK-PD predictions corrected by neural networks is denotes as $\hat{Y}_{i} = [\hat{y}_{i}(\tau-1), \cdots, \hat{y}_{i} (\tau-l)]$, the ground true history of BIS is denotes as $Y_{i} = [y_i(\tau-1), \cdots, y_i (\tau-l)], i \in N$, $N$ is training batch size. The final predicted BIS and ground true is denotes as $\hat{y_i}$ and $y_i$.

To provide the network with a preliminary BIS history, $L_{h}\left(\hat{Y}, Y; \theta_{e n}\right)$ is adopted in the encoder part such that the encoder extracts a BIS value that is as close as possible to the historical true value trend:
\begin{equation}
L_{h}\left(\hat{Y}, Y ; \theta_{en}\right)=\frac{1}{NT}\sum_{i=1}^{N}\sum_{t=1}^{T}\left(\hat{y}_{i}(\tau-t)-y_{i}(\tau-t)\right)^{2}
\end{equation}

Based on the length of the drug in fusion history, we set $T=180$ in the experiments, which corresponds to the length of the input sequence.

In the outset, we used the standard mean squared error (MSE) loss function as the objective function for gradient descent to train the model:
\begin{equation}
L_{MSE}\left(\hat{y}, y ; \theta_{en}, \theta_{fl}, \theta_{de}\right)=\frac{1}{N}\sum_{i=1}^{N}\left(\hat{y}_{i}-y_{i}\right)^{2}
\end{equation}

However, we found that using only the MSE loss function does not encourage the model to learn the mutation condition of BIS, particularly during the maintenance period of anesthesia (i.e., from 10 minutes after anesthetic injection to the end of drug infusion), where the model tends to learn the mean of the sample. Therefore the model can be further improved by introducing the reductive bias, a weighted combination of multiple loss functions, with each particular function that is used to focus on a different side. Therefore, we use a weighted MSE loss as follows:
\begin{equation}
L_{w}\left(\hat{y}, y ; \theta_{en}, \theta_{fl}, \theta_{de}\right)=\frac{1}{N}\sum_{i=1}^{N}w_i\left(\hat{y}_{i}-y_{i}\right)^{2}, w_i \in W
\end{equation}
\begin{equation}
W=\frac{1}{\tilde{p}(y^{\prime})}=\frac{1}{\int_{y} k\left(y, y^{\prime}\right) p(y) d y}
\end{equation}
where the weight matrix \(W=[w_{1},  w_{2}, \cdots, w_{100}]\) is the inverse of the new label distribution \(  \tilde{p}(y^{\prime})  \) after kernel smoothing. The smaller the number of samples, the larger the weights \( w_{i}\). Correspondingly, the loss of the sample is larger.
Overall, the optimal model can be obtained by:

\begin{equation}
\begin{gathered}
\arg \min (
\lambda_{h} L_{h}\left(\hat{Y}, Y; \theta_{en}\right) 
+\lambda_{w}L_{w}\left(\hat{y}, y ; \theta_{en}, \theta_{fl}, \theta_{de}\right))\\
\end{gathered}
\end{equation}
where \(\lambda_{h}\) and \(\lambda_{w}\) are the loss weights. In the experiments, we set them to 5 and 10 respectively.

\section{Experiments and Result}

\begin{table*}[t]
\renewcommand\arraystretch{2}
\caption{Comparison of errors between ours proposed model and the baseline model during three anesthesia periods}
\begin{center}
\resizebox{\textwidth}{22mm}{
\begin{tabular}{c ccc ccc ccc}
\hline
&
\multicolumn{3}{c}{MDPE(\%)}&
\multicolumn{3}{c}{MDAPE(\%)}&
\multicolumn{3}{c}{RMSE} \\
\cmidrule(r){2-4} \cmidrule(r){5-7} \cmidrule(r){8-10}
Anesthesia Period &
\multicolumn{1}{c}{PK-PD}& LSTM & Ours &
\multicolumn{1}{c}{PK-PD}& LSTM & Ours &
\multicolumn{1}{c}{PK-PD}& LSTM & Ours \\

\hline

All&
\multicolumn{1}{c}{21.75 ± 12.65} & 3.64 ± 14.96 & \textbf{-2.08} ± 14.91  &
\multicolumn{1}{c}{24.23 ± 10.16} &15.97 ± 7.91  &\textbf{15.51} ± 6.87  &
\multicolumn{1}{c}{15.64 ± 5.19} &10.20 ± 2.45  & \textbf{9.52} ± 2.35                  \\

Induction         &
\multicolumn{1}{c}{-16.09 ± 23.12} &-6.35 ± 20.50  & \textbf{4.64} ± 17.82  & \multicolumn{1}{c}{27.18 ± 13.89} &22.75 ± 9.39  & \textbf{18.52} ± 8.32  & \multicolumn{1}{c}{17.62 ± 4.56} &14.57 ± 3.79 & \textbf{12.91} ± 4.14                  \\

Maintenance       &
\multicolumn{1}{c}{22.89 ± 12.74} &-2.99 ± 15.24  & \textbf{-2.62} ± 15.32  & \multicolumn{1}{c}{24.34 ± 10.70} &\textbf{15.08} ± 8.34  & 15.26 ± 7.51  & \multicolumn{1}{c}{14.78 ± 5.84} &8.72 ± 2.78  & \textbf{8.50} ± 2.62              
\\
Recovery          & 
\multicolumn{1}{c}{16.54 ± 18.15} &-13.92 ± 25.44 & \textbf{-4.00} ± 22.23 & \multicolumn{1}{c}{22.29 ± 12.46} &24.97 ± 16.03 & \textbf{19.18} ± 13.64 & \multicolumn{1}{c}{16.93 ± 7.86} &15.18 ± 6.43 & \textbf{12.38} ± 5.55           
\\ 
\hline
\end{tabular}}
\end{center}
\label{vitaldb-result}
\end{table*}

\subsection{Data Preparation}
In our experiments, the used data is the VitalDB database, which includes the drug injection records and static covariates of patient physiological characteristics (age, gender, height, weight). The detailed description is illustrated in Section B of Methodology.
Since the VitalDB database contains the real surgery records collected in real time, there is a lot of noise, interference, and incorrect records in the data, which greatly affect the learning of the model. Additional data processing is required to convert it into a suitable form for computer computation. Therefore, the database is cleaned to minimize the interference of the noisy signals.

Considered that a large amount of noise and error records is in the database, we first performed data cleaning. Since there are many missing values in some samples, the following operations are performed: 1) interpolate the data outliers and nulls with linear interpolation; 2) discard samples with more than 30s data loss; 3) discard samples with only half-field surgery records. After that, the data are subjected to additional processing. Considering the retention time of the drug on the human body, propofol clears 66\% of the time after 3 hours of injection for about 25 minutes \cite{struys2000comparison}, we set the time range of the drug input history within 1800s to extract the features, which is consistent with the hypothesis in \cite{hochreiter1997long}. 
It is worth noting that during the initial periods, the input, in the form of a zero sequence concatenate a medication record, is used for our proposed model to predict the BIS value. That means, when $t \in [1, 1800] s$, the input is a zero sequence of length $(1800-t)$ seconds concatenate a $t$ seconds of medication record.

The injection histories of propofol and remifentanil in the original database are the total drug injections recorded every 1s. To save the computing resources, the data are processed into the drug doses injected in every 10s (i.e., 180 data points). In addition, the drug dose is divided by the length of time to obtain the drug injection rate, which often is used in the PK-PD model to calculate BIS characteristics. The drug injection history and other static covariates are normalized to facilitate faster network convergence.

\begin{figure*}[htbp]
\centering
\includegraphics[scale=0.62]{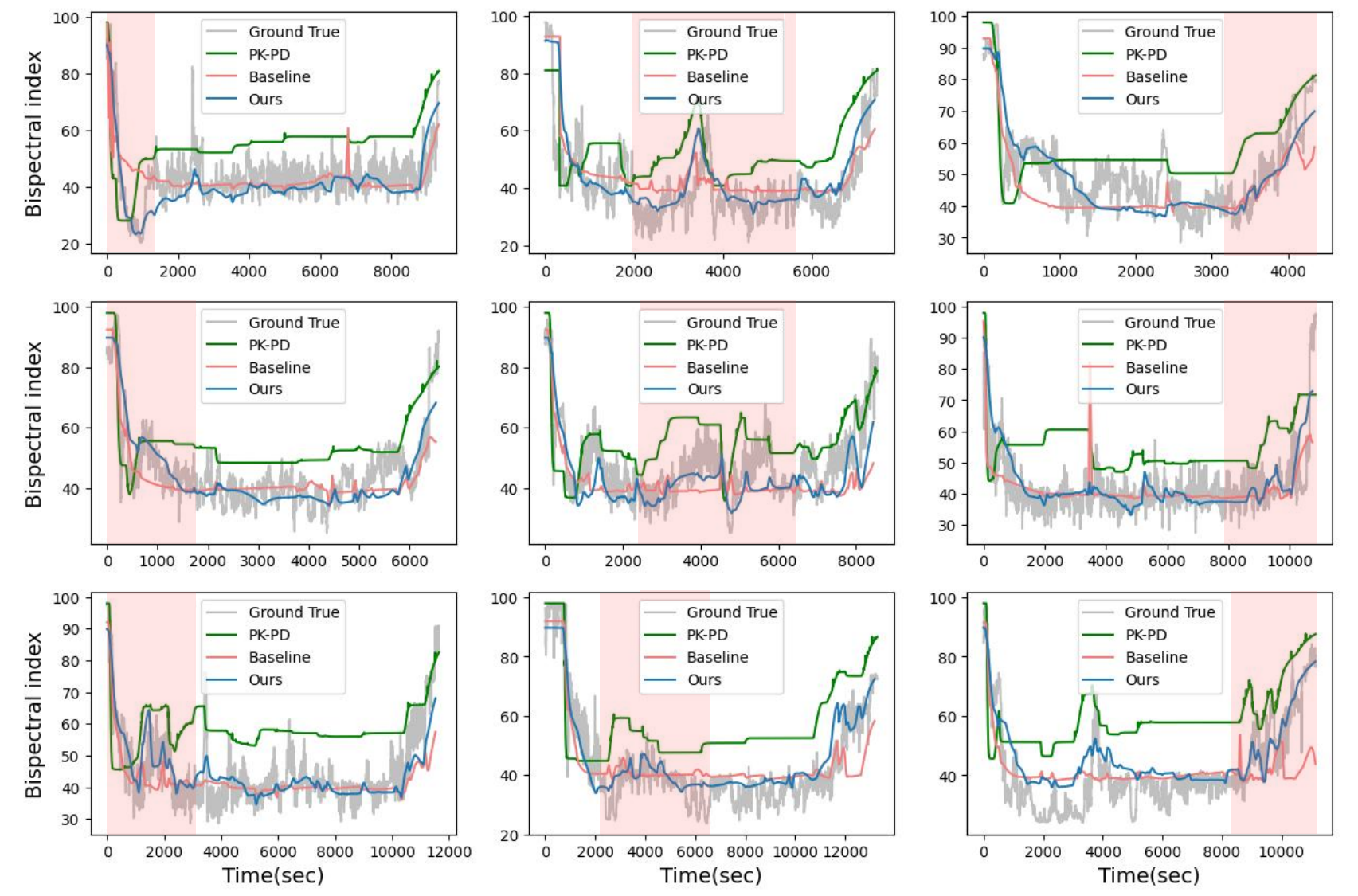}
\caption{Performance comparison between our proposed method and other compared methods (i.e., the baseline LSTM method \cite{lee2018prediction} and the PK-PD \cite{short2016refining} method)
on different test samples. Each subfigure corresponds to an independent sample.}
\label{compare baseline and ours}
\end{figure*}


To suppress the negative influence of the noise contained in the database, the true BIS values of the training set is smoothed and the locally weighted scatter plot smoothing (LOWESS) with a smoothing parameter of 0.03 is used for the original BIS values to reduce the computational error during the training phase. To ensure the authenticity of the experiments, the validation set and test set are not smoothed.
\subsection{Data Characteristics}

We use the VitalDB database as our capital training set, which is a open access data source, freely downloadable from the website, \href{https://vitaldb.net}{https://vitaldb.net}. 
In our experiments, 680 samples that contain the real surgical records with TIVA general anesthesia injection are randomly selected from the VitalDB database as the original database. After data cleaning, 348 samples with serious missing data (e.g., only half of the surgical records) are excluded. In this case, only 332 samples are used as the remaining samples. Among them, 180 samples are randomly selected as the training set, 76 samples are randomly selected as the validation set, and the other samples are used as the test set. The characteristics of these three sets of samples are shown in the  Table \ref{data stastic}.



\subsection{Experimental Settings}

The evaluation metrics including median performance error (MDPE), median absolute performance (MDAPE) and mean square error (RMSE) are used to evaluate the performance of our proposed model.
The predictive performance is evaluated for each period according to the definition of three periods in anesthesia surgery (induction period: from the start of the anesthetic drug propofol injection until 10 minutes later, maintenance period: from the end of the induction period until the cessation of prorofol injection, and recovery period: from the cessation of propofol injection until the end of the surgical record). In addition, a paired t-test is used to compare the performance of our model with other compared methods, and the experimental results are expressed as mean ± SD (range). Statistical analysis is performed with SPSS 21 (IBM, USA), and $P < 0.05$ is the considered significant for the paired t-test.

The model is optimized by using the Adam optimizer. Batchsize is set to 1024 while training, the initial learning rate is 0.03, and the learning rate decays to 0.1 times after every 10 epochs.
Unless otherwise mentioned, the parameters of PK-DK model used in our proposed method are shown in Table \ref{pk-pd params}.

The pytorch 1.4.0 framework and python 3.9.1 are adopted in our implementations. All experiments are run on a single 24GB NVIDIA TITAN RTX GPU, which takes about 30 minutes to train 50 epochs with batch size 1024 on the entire dataset. Our code is made available at \href{https://github.com/heeeyk/Transformer-DOA-Prediction}{https://github.com/heeeyk/Transformer-DOA-Prediction}.
\subsection{Experimental Results}
In the experiments, our proposed model has made a comparison with the LSTM method \cite{lee2018prediction} and the PK-PD method\cite{short2016refining}, the experimental results are shown in the Table \ref{vitaldb-result}. From this table, one can observe that the proposed model outperforms the baseline method and the PK-PD method in all periods in terms of evaluation metrics except for MDAPE in the maintenance period, in which our proposed model has a slight performance degradation.
The main reason is that the baseline method tends to predict the smooth BIS curves, which coincides with the overall trend in the maintenance period with a very large sample size, and thus has a better performance on the whole dataset. The detailed analysis is described in the next section. The performance comparison on different test samples is shown in Fig. \ref{compare baseline and ours}. Obviously, our method has greatly improvements on the predictive capability for the mutation conditions and outperforms the baseline method, which are performed significantly in the light pink areas. In addition, the concordance correlation coefficient (CCC) is used to measure the correlation between BIS and ground true BIS. The experimental results for all methods are shown in Table \ref{CCC}. The CCC (95\% Confidence Interval) is 0.677 [0.691 to 0.665] in our model, which is significantly larger than that in the LSTM method (0.590 [0.582 to 0.609]) and in the PK-PD method (0.556 [0.543 to 0.571]).

In the real-world applications, the DOA prediction is often set to predict the BIS value every 1 second. Therefore, we set the batch size to 1 to simulate the inference process of the DOA prediction, and find that our model only takes 0.014 seconds to predict the BIS value every single time. This indicates that our model can be used for real-time monitoring.


\begin{figure*}[htbp]
\centering
\subfigure[]
{
    \begin{minipage}[b]{.3\linewidth}
        \centering
        \includegraphics[scale=0.3]{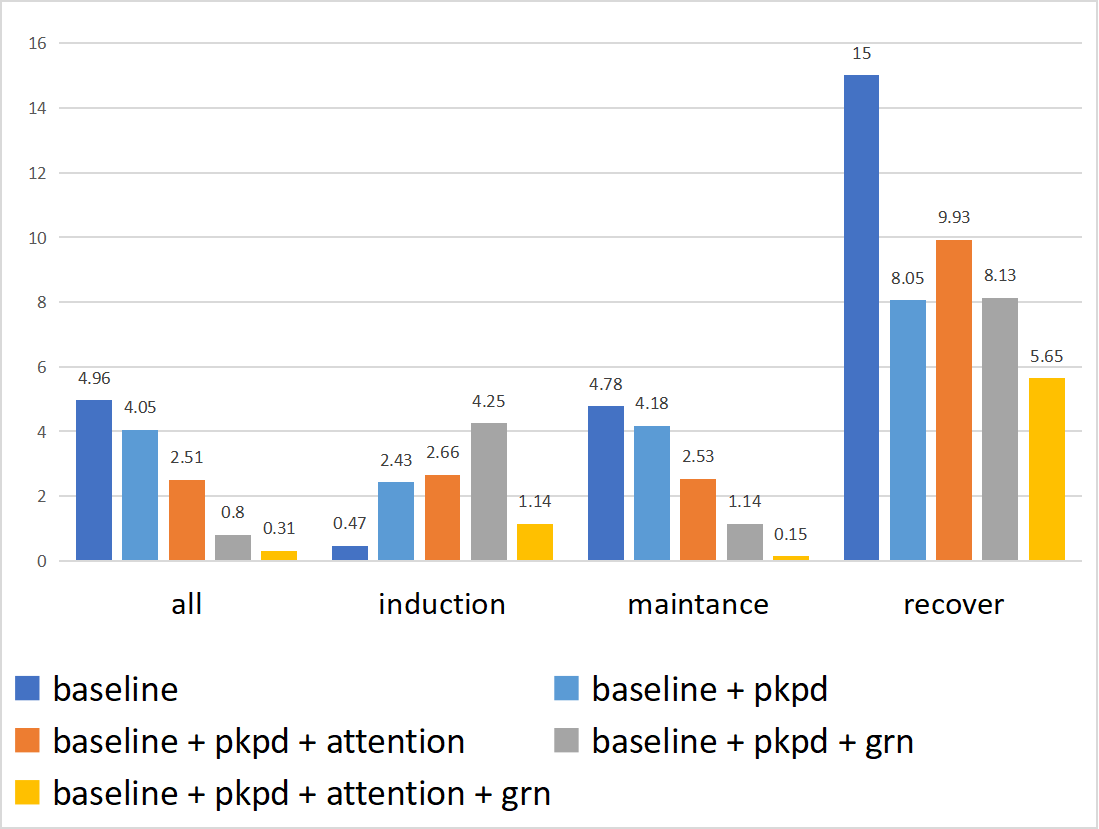}
    \end{minipage}
   }
\subfigure[]
{
 	\begin{minipage}[b]{.3\linewidth}
        \centering
        \includegraphics[scale=0.3]{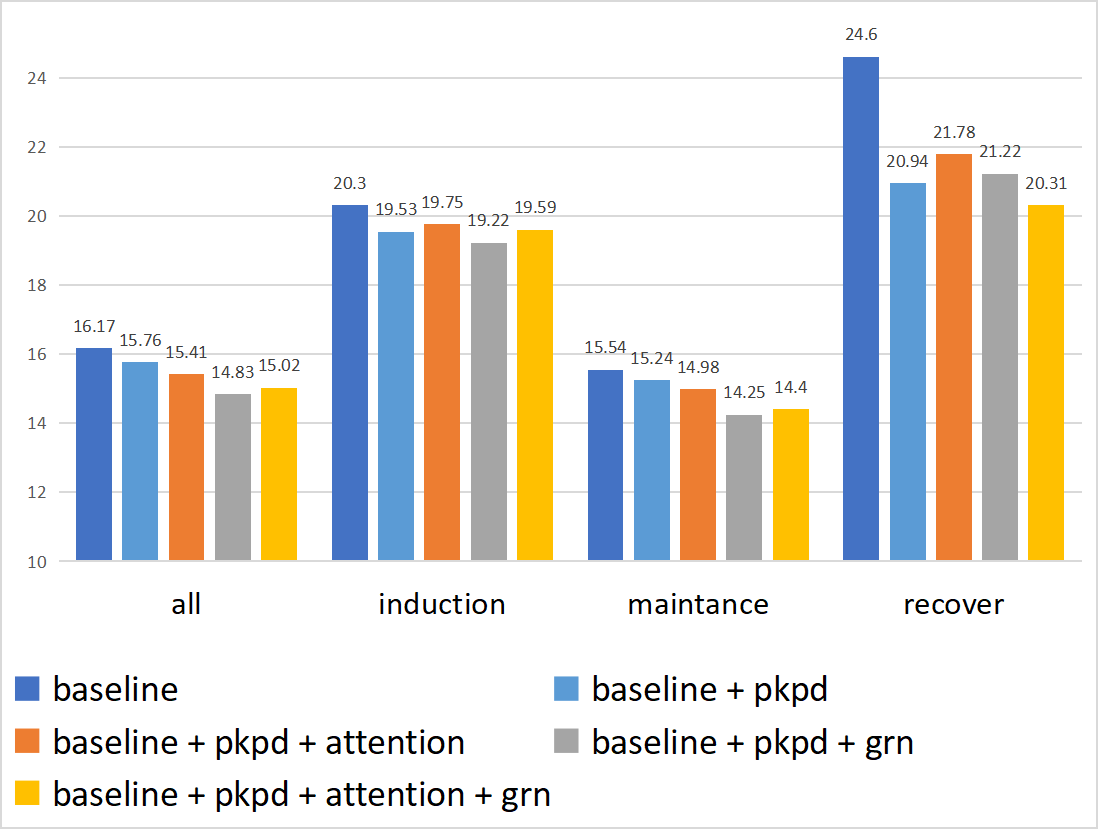}
    \end{minipage}
}
\subfigure[]
{
 	\begin{minipage}[b]{.3\linewidth}
        \centering
        \includegraphics[scale=0.3]{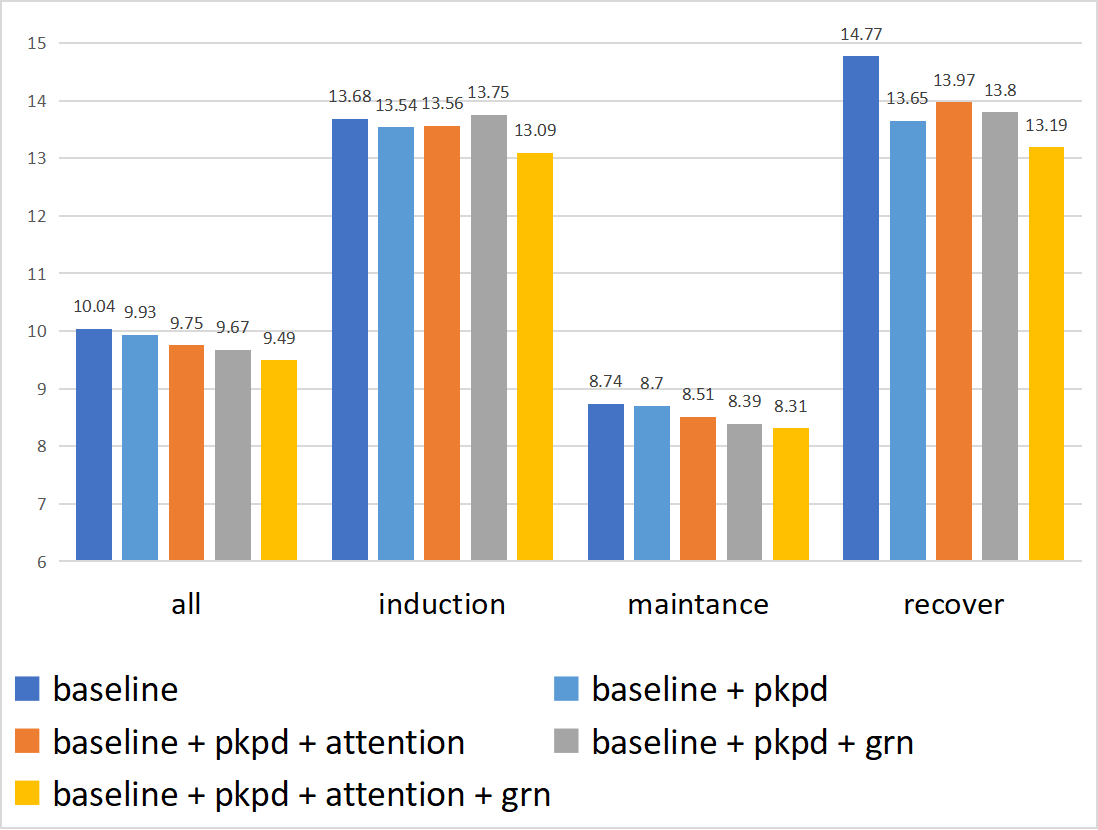}
    \end{minipage}
}
\caption{Prediction results of ablation study on different components for our proposed model. (a): MDPE; (b): MDAPE; (c): RMSE.}
\label{ablation}
\end{figure*}

\begin{table}
\centering
\caption{CCC (concordance correlation coefficient) comparisons}
\renewcommand\arraystretch{2}
\setlength{\tabcolsep}{4mm}{
\begin{tabular}{lccc}

\hline
&PK-PD    & LSTM   &  Ours \\
\hline
CCC & 
$0.556_{0.543}^{0.571}$ & $0.595_{0.582}^{0.609}$ &  $0.677_{0.665}^{0.691}$ \\
\hline
\multicolumn{4}{l}{\small  \makecell[l]{$\bullet$
$C_{b}^{a}$ denotes concordance correlation coefficient \\ with  95\%  confidence interval upper bound $a$ and \\ lower bound $b$.}}
\end{tabular}}
\label{CCC}
\end{table}


\begin{figure}[htbp]
\centering
\includegraphics[width=3.4in, height=2.6in]{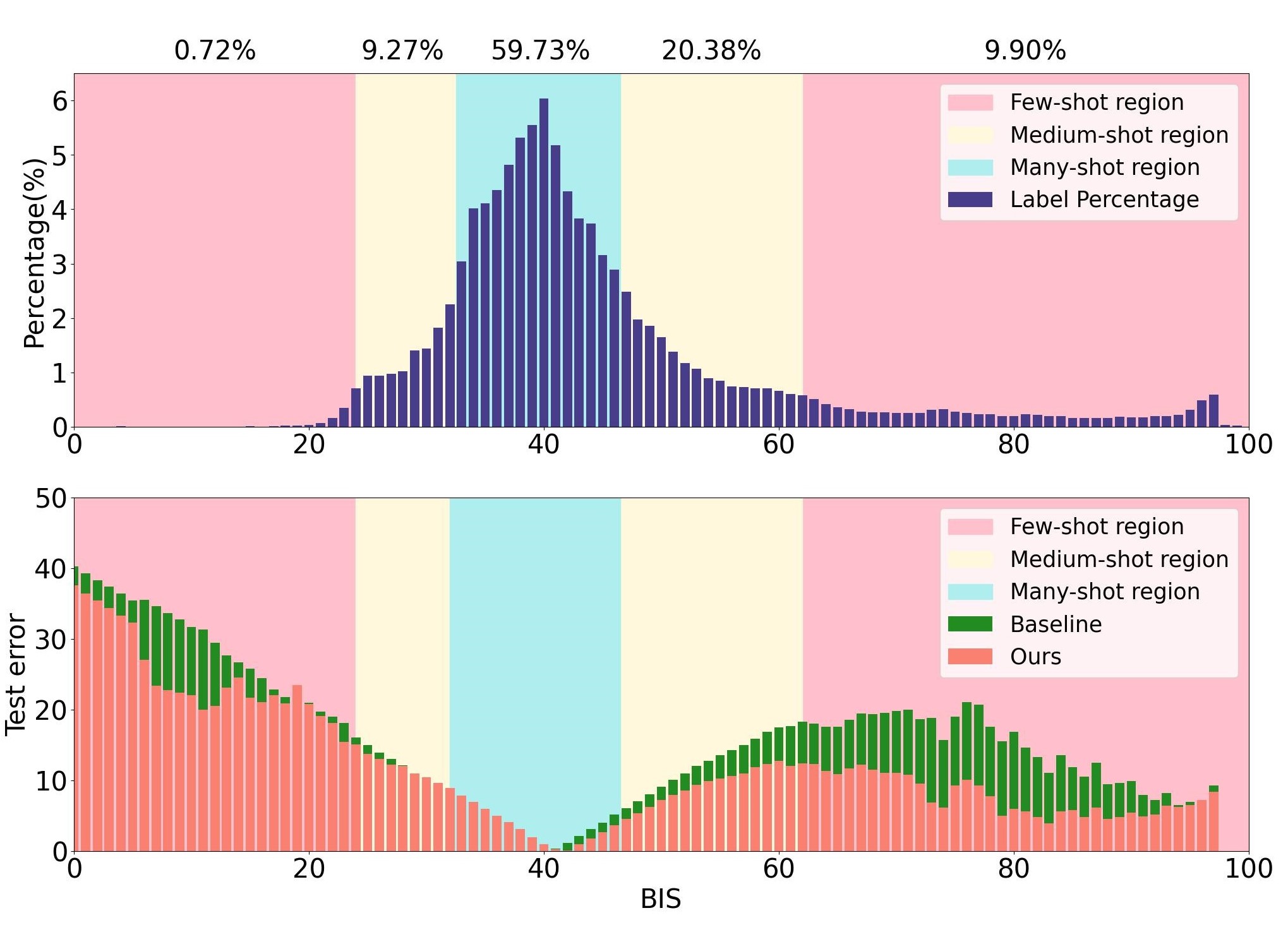}\vskip -0.3cm
\caption{Top: the sample label distribution of the dataset is divided into three regions, in which the largest amount of data in the many-shot region contains 59.73\% and the smallest amount of data in the few-shot region contains 10.62\%. Bottom: comparison of the baseline method with our proposed method on the test error.}
\label{test error}
\end{figure}

\subsection{Test Error Analysis}
For a single case, one can observe the extreme unreasonableness of prediction curve for the baseline method, such as lacking variation and fluctuating in a rough range around 40 for the BIS values. This coincides with the label distribution of the dataset, as shown in the upper part of Fig. \ref{test error}.
By analyzing the label distribution of the dataset, it is verified that the baseline method is disappointing in terms of the overall prediction performance, because it has a overfitting problem in the many-shot region with large data volume and neglects the prediction ability for other regions, especially the unbalanced data distribution happens to be the medical data.

Therefore, the testing errors in the sample labels is adopted to verified the fact that our method outperforms the baseline method. Specifically, the test error is calculated by the following formula:
\begin{equation}
error^{(j)}=\frac{1}{n^{(j)}} \mid \sum_{i=1}^{l}\left([\hat{Y}_{i}]^{(j)}-[Y_{i}]^{(j)}\right)\mid, j \in[0,100]
\end{equation}
where \(j\) denotes the range of the BIS values, \(n^{(j)}\) is the number of points in the dataset, \([.]\) stands for the rounding operation, \(\hat{Y}_{i}\) and \({Y}_{i}\) denote the predicted output of the model and the ground true for the \(i^{th}\) sample, respectively.
As shown in Fig. \ref{test error}, one can see that, compared with the baseline method, our proposed method has great improvements on the prediction performance in other regions, and at the same time, without reducing the predictive power in the many-shot region. Especially in the few-shot interval between 10 to 20 and between 60 to 90, the test error decreases by \textbf{15.23\%} and \textbf{48.99\%}, respectively.
These experimental results are coincide with the rare deep anesthesia state and the shallow anesthesia state between wakefulness and anesthesia during the induction period.
Moreover, the enhancement of the predictive ability over these two regions indicates that our proposed method solves the overfitting problem in the many-shot region.

\begin{figure*}[htbp]
\centering
\includegraphics[scale=0.56]{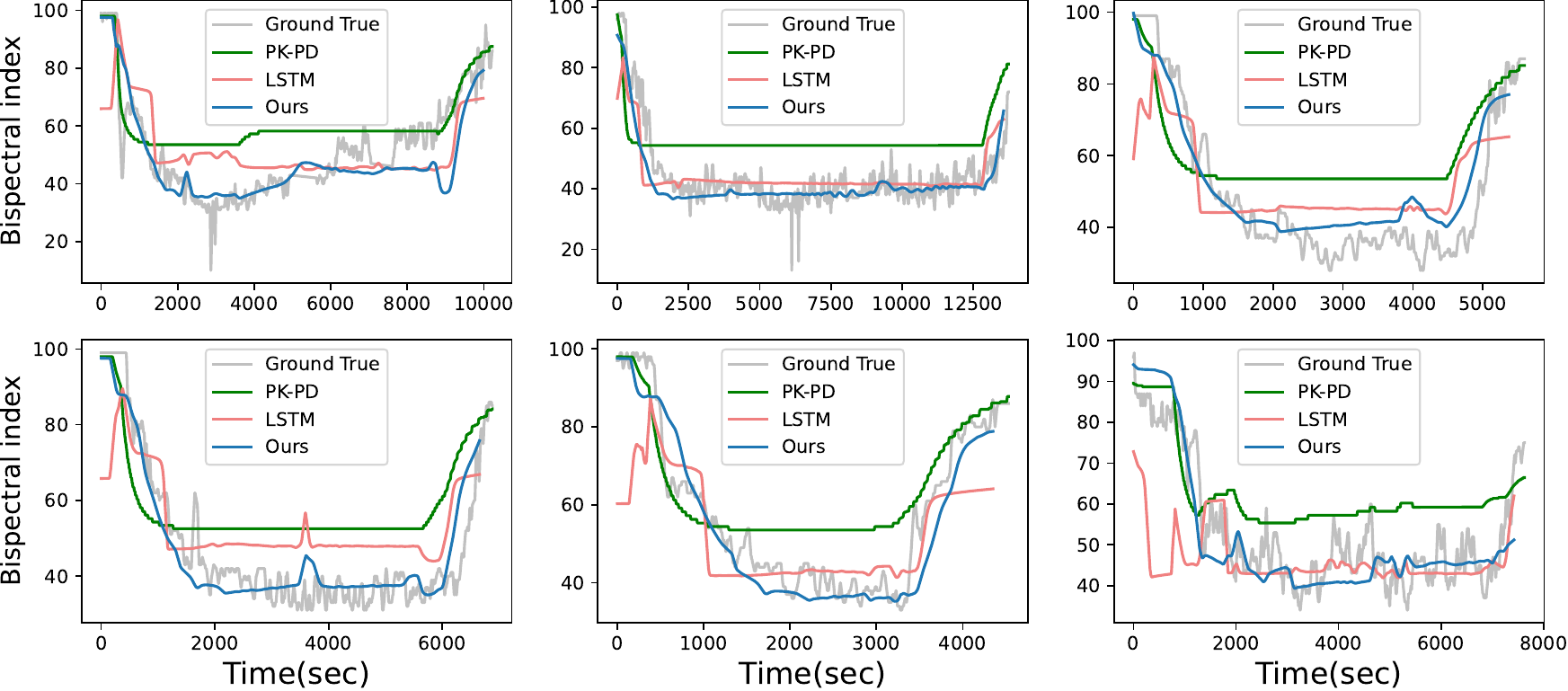}
\caption{Performance comparison between our proposed method and the baseline methods (LSTM\cite{lee2018prediction} and PK-PD\cite{short2016refining}) on our in-house dataset.}
\label{ni compare}
\end{figure*}


\subsection{The Region Where BIS Mutations Occur}
For all experimental data, the main mutations of BIS occurs in the medium-shot region, in which the number of the samples is less than 30\%. In order to improve the prediction performance, a weighted MSE loss function is adopted in the proposed model to solve the issue of data imbalance, such that much attention is paid to the medium-shot region.
We have conducted a statistical analysis during the relatively stable anesthesia maintenance phase, and the percentage of the mutations of BIS in different regions is shown in Table \ref{percentage of the mutations} From this table, one can see that, if the BIS value, denoted as $B_t$ at time $t$, satisfies one of the following conditions:
\begin{equation}
\begin{cases}
\left | B_t-min(B_T) \right | >m 
 \\
\left | B_t-max(B_T) \right | >m
\end{cases}
,\space  T\in(t-30, t+30)s
\end{equation}
where $m$ denotes the magnitude of BIS changes, and a larger $m$ indicates a greater mutation magnitude. In Table \ref{percentage of the mutations}, we have illustrated the results with $m=5, 7$ and $10$ respectively. In particular, when $m=10$, approximately 33\% of the BIS mutations occurred in the Many-shot region ($BIS\in\left[31,48\right]$). However, the rest of the BIS mutations occurred in the Medium-shot and Few-shot regions. This means, the loss function assigns the weight to these data points in the Medium-shot and Few-shot regions more than twice the weight in the Many-shot region. Therefore, our loss function enables the network to pay more attention to the mutation condition of BIS. This indicates, the number of samples and the mutation condition of BIS has necessary relationship in our proposed model.




\begin{table}
\renewcommand\arraystretch{2.5}
\centering
\caption{Percentage of the mutations of BIS in different regions}
\begin{tabular}{ccccc}
\hline
Region
&$m=5$    
&$m=7$    
&$m=10$   
& \makecell[c]{The weights of \\ loss function} \\
\hline
\makecell[c]{Many-shot \\ $BIS\in[31,\ 48)$ }
& 50.94\% & 34.82\% & 33.33\% &$ w\in[1,\ 2)$  \\

\makecell[c]{Medium-shot \\ $BIS\in[48,\ 54)$ }
& 21.23\% & 27.93\% & 33.33\% &$ w\in[2,\ 4)$\\

\makecell[c]{Medium-shot \\ $BIS\in[54,\ 64)$}    
& 16.86\%  & 26.89\% & 19.04\% &$ w\in[4,\ 8)$  \\

\makecell[c]{Few-shot \\ $BIS \ge 64$} 
& 7.46\%  & 10.34\% & 14.28\% &$ w \ge 8$ \\
\hline
\end{tabular}
\label{percentage of the mutations}
\end{table}

\subsection{Ablation Experiments}
To illustrate the contribution of each individual module in our proposed model, the ablation experiments are conducted. Starting from our full network model and gradually removing some of the network components, each method is trained under the same conditions until the network converged, and the experimental results in terms of MDPE, MDAPE and RMSE are shown in Fig. \ref{ablation} respectively.
From this figure, one can observe that, each network component has a contribution to improve the performance metrics (e.g., MDAPE and RMSE) for the prediction of DOA.
Our proposed model outperforms the baseline method in the entire anesthesia period. However, the prediction performance may degrade  during the induction period. Based on the experimental results, this problem may be caused by the introduction of the PK-PD model, because the PK-PD model in the induction period predicts the BIS values with large deviations (e.g., there is a certain time lag). This will lead to the performance degradation. In addition, if the GRN component is discarded and only the attention layer is considered, the effect will not be greatly improved compared with the combination of the baseline method plus PK-PD. The main reason is that the lack of the GRN component prevents the effective inclusion of static covariates in the temporal information, which often leads to a bad prediction performance.

\begin{table*}[t]
\renewcommand\arraystretch{2}

\caption{Comparison of errors between ours proposed model and other methods in our in-house dataset}
\begin{center}
\resizebox{\textwidth}{24mm}{
\begin{tabular}{c ccc ccc ccc}
\hline
&
 \multicolumn{3}{c}{MDPE(\%)}&
 \multicolumn{3}{c}{MDAPE(\%)}&
 \multicolumn{3}{c}{RMSE} \\
\cmidrule(r){2-4} \cmidrule(r){5-7} \cmidrule(r){8-10}
Anesthesia Period &
\multicolumn{1}{c}{PK-PD}& LSTM & Ours& \multicolumn{1}{c}{PK-PD}& LSTM &  Ours& \multicolumn{1}{c}{PK-PD}& LSTM &  Ours \\

\hline

All&
\multicolumn{1}{c}{24.54±16.62} & 3.79±20.55 & \textbf{2.83}±19.46  &

\multicolumn{1}{c}{28.1±13.24}  & 20.83±11.18 & \textbf{20.08}±9.40  &

\multicolumn{1}{c}{17.94±6.27}  & 13.23±3.88 & \textbf{12.08}±3.73                 \\

Induction        
& \multicolumn{1}{c}{-19.18±15.63}  & -8.35±20.77 & \textbf{-0.46}±13.39  &

\multicolumn{1}{c}{23.81±9.94}  & 20.35±12.10& \textbf{12.91}±6.95  &

\multicolumn{1}{c}{16.52±4.80} & 16.38±6.21 & \textbf{12.86}±5.47                  \\

Maintenance      &
\multicolumn{1}{c}{26.99±16.69}  & 4.71±21.89 & \textbf{2.55}±21.06  &

\multicolumn{1}{c}{28.92±14.19}  & \textbf{20.76}±12.28 & 20.87±10.19 &

\multicolumn{1}{c}{17.45±7.09}  & 11.78±4.42 & \textbf{10.95}±4.25                   \\

Recovery         &
\multicolumn{1}{c}{20.27±21.10} & -2.85±28.93 & \textbf{2.37}±27.45 &

\multicolumn{1}{c}{24.39±16.55} & 27.22±15.10 & \textbf{24.32}±13.56 &

\multicolumn{1}{c}{19.09±9.79} & 15.98±6.51 & \textbf{15.08}±6.50                    \\ \hline
\end{tabular}}
\end{center}
\label{ni results}
\end{table*}

In general, incorporating the PK-PD model into the network can improve the prediction performance of the model because it contains a large number of hyperparameters calculated from human experiments itself. Furthermore, it provides some statistical data to the deep learning model for enhancing the robustness of the model. 
The PK-PD model also does regression calculations on the past BIS indices, which brings much time series information to the model. But the disadvantage is that the PK-PD model has bias in the induction period, which easily leads to partial performance degradation of the model.
To analyze the contribution of the PK-PD model in our method, the combination of the baseline (i.e., LSTM) model and the PK-PD model and the baseline model are conducted in the ablation experiment. From the experimental result in Fig. \ref{ablation}(a), one can observe that the performance of the combination of them can improve in most periods, except for the induction period. In particular, the MDPE in the induction period increases from 0.47 (only LSTM) to 2.43 (LSTM+PK-PD), which degrades the performance of  the model. However, when the other modules (i.e., Attention and GRN) are incorporated into our model, the MDPE in the induction period drops to 1.14, which illustrates our model can reduce the negative influence efficiently for the disadvantage of the PK-PD model.

The GRN module, on the other hand, can favourably reflect the physiological changes of drug doses in different populations by adding the static information to the time series (different ages have different drug elimination rates), and the unique gating mechanism of GRN can eliminate the signal noise and the static variables that have almost no contribution to the output. The attention mechanism can learn the long-term relationship between different time steps, but it is difficult to perform effectively when the dynamic-static information is missing in GRN. The confounding effect of temporal information and static covariance is considered to maximize the effect of the attention mechanism.

\begin{table}
\renewcommand\arraystretch{1.6}
\centering
\caption{Datasets difference}
\setlength{\tabcolsep}{0.7mm}{
\begin{tabular}{lcc}
\hline
& Our dataset          
& VitalDB(Training set) \\
\hline
N 
& 44 & 180   \\
Age(yr) 
& 39.9 ± 13.4 (19-69)  & 56.1 ± 14.0 (17-82)\\
Sex(male/female)               
& 22/22 & 113/67 \\
weight(kg)  
& 62.6 ± 10.5 (43-105) & 61.5 ± 10.2 (37.9-98.1)  \\
Height(cm) 
& 166 ± 8.1 (147-183)  & 163.2 ± 8.2 (138.8-186.6) \\
Median BIS
&   43.5 ± 10.1 (23.0-68.2)   &  41.1 ± 5.4 (25.9-59.9)    \\
\hline
\end{tabular}}
\label{in-house dataset}
\end{table}

\subsection{Model Generalization}
To verify the generalization of our model, the experiments that training a model on the VitalDB source and testing it on our in-house dataset are conducted. Our in-house dataset contains 44 cases collected from a hospital in China. 
The main difference between the the VitalDB dataset and our dataset is shown in Table \ref{in-house dataset}. 
Since the two datasets are collected from different race and data acquisition equipment, it may lead to the large domain gap. In particular, our dataset is mainly sampled from younger people, hence the average age of samples in our dataset is the larger that that of samples in the VitalDB dataset.
In the experiments, 20 cases are randomly selected as the training set to fine-tune the baseline and our model. 
The PK-PD cannot be fine-tune, since its parameters is fixed.
The experimental results are shown in Fig. \ref{ni compare} and Table \ref{ni results}, and obviously our method still has the best performance. It's worth noting that, due to the limitation of our collection equipment, we can only record the BIS values every 5 seconds. In this situation, the BIS and narcotic record may cause inaccurate label information after interpolation.

\section{Conclusion}
Accurate drug efficacy prediction is helpful for anesthesiologists to make suitable decisions in the clinical procedures. In this paper, a transformer-based prediction method is proposed for predicting the depth of anesthesia. Particularly, the proposed method adopts a LSTM based deep learning architecture and an enhanced attention mechanism to efficiently predict the sudden change of anesthesia depth under the effect of drugs. In addition, a weighted loss function is used in the network to solve the problem of data imbalance, improving the generalization in comparison to previous approaches. Experimental results show that our proposed model has better prediction performance than previous methods, especially in the few-shot region such as deep anesthesia stage and situation.

\bibliographystyle{unsrt}
\bibliography{references.bib}

\end{document}